\documentclass[
]{ceurart}

\usepackage{listings}
\begin{document}

\copyrightyear{2023}
\copyrightclause{Copyright for this paper by its authors.
  Use permitted under Creative Commons License Attribution 4.0
  International (CC BY 4.0).}

\conference{CLEF 2023: Conference and Labs of the Evaluation Forum, September 18–21, 2023, Thessaloniki, Greece}

\title{Keeping in Time: Adding Temporal Context to Sentiment Analysis Models}

\title[mode=sub]{Notebook for the LongEval Lab at CLEF 2023}


\author[1]{Dean Ninalga}[%
orcid=0009-0008-4246-1936,
email=djninalga@gmail.com
]
\cormark[1]
\address[1]{Toronto, Canada}

\cortext[1]{Corresponding author.}

\begin{abstract}
This paper presents a state-of-the-art solution to the LongEval CLEF 2023 Lab Task 2: \emph{LongEval-Classification} \cite{clef-longeval-alkhalifa-2023}. 
The goal of this task is to improve and preserve the performance of sentiment analysis models across shorter and longer time periods.
Our framework feeds \emph{date-prefixed} textual inputs to a pre-trained language model, where the timestamp is included in the text. We show \emph{date-prefixed} samples better conditions model outputs on the temporal context of the respective texts.
Moreover, we further boost performance by performing self-labeling on unlabeled data to train a student model. We augment the self-labeling process using a novel augmentation strategy leveraging the \emph{date-prefixed} formatting of our samples.
We demonstrate concrete performance gains on the LongEval-Classification \cite{clef-longeval-alkhalifa-2023} evaluation set over non-augmented self-labeling.
Our framework achieves a 2nd place ranking with an overall score of 0.6923 and reports the best \emph{Relative Performance Drop} (RPD) \cite{Alkhalifa2021OpinionsAM} of -0.0656 over the short evaluation set (see \citet{longevaloverview2023}).
\end{abstract}

\begin{keywords}
  Self-Labeling \sep
  Sentiment Analysis \sep
  Temporal Misalignment \sep
  Date-Prefixing
\end{keywords}

\maketitle

\section{Introduction}

The application of language models such as BERT \cite{Devlin2019BERTPO}, RoBERTa \cite{Liu2019RoBERTaAR} and XLM-RoBERTa \cite{Conneau2019UnsupervisedCR} to textual data is a core component in many natural language processing (NLP) pipelines. 
However, a notable limitation of most language models is their lack of temporal awareness, as they typically encode text into fixed representations. Conversely, the nature of textual data is inherently dynamic and subject to change over time. Where traditional meanings of words, phrases, and concepts are constantly evolving \cite{Dhingra2021TimeAwareLM, Margatina2023DynamicBO}.
Furthermore, significant events can alter the factual basis of the text \cite{Agarwal2021TemporalEO}.
Although metadata of well-known text corpora includes timestamps, timestamps are almost never used within many NLP pipelines. A sentiment analysis model trained today could interpret the phrase: "You are just like X" as a positive sentiment. However, an issue can arise once people consider a comparison to `X’ as a non-positive comparison. Subsequently, the model becomes \emph{misaligned} if this flip in public opinion occurs. Hence, it can be difficult to train models that can generalize to future data without a sense of temporal context and awareness \cite{Lazaridou2021MindTG}. 

Mitigating \emph{temporal misalignment} \cite{Luu2021TimeWF} between the facts and general sentiments of the current world and those found in text corpora is an active area of focus in various areas of research in nlp. 
In particular, work in NER (named-entity-recognition) \cite{Pustejovsky2005TemporalAE, Agarwal2021TemporalEO, Zhang2023MitigatingTM} and question-and-answering \cite{Jang2021TowardsCK, Chen2017ReadingWT, Pustejovsky2005TemporalAE, Livska2022StreamingQAAB} often directly address temporal misalignment as they are considered \emph{knowledge-intensive} tasks \cite{Lazaridou2021MindTG}. 

A common and straightforward way to address temporal misalignment in textual data is to create new models (or update old ones) with the most recent data available \cite{Loureiro2022TimeLMsDL, Jang2022TemporalWikiAL, Lazaridou2021MindTG}.
However, continually growing datasets incur an increase in computational costs for data acquisition and training models which also contributes to an ever-increasing environmental cost \cite{Strubell2019EnergyAP, Attanasio2022IsIW}.
Therefore, finding a solution outside of continuous retraining that preserves model performance over time is desirable.

In this paper, we follow \citet{Dhingra2021TimeAwareLM} who use an alternative approach that modifies the textual input with its timestamp. Thus, we can take advantage of text-only pre-trained language models used for classification in addition to conditioning the models with the temporal context for the input.

We will outline our system, which is aligned with some of the recent works in NER and temporal misalignment, and evaluate it on the \emph{LongEval-Classification} benchmark \cite{clef-longeval-alkhalifa-2023}.

Our contribution is two-fold:
(1) We show that date-prefixing the input text with its timestamp conditions the outputs of a language model on the temporal context of the input.
(2) We utilize an augmentation strategy that leverages the date-prefixing by randomly modifying the timestamp of unlabeled inputs. We show that this augmentation strategy improves the performance benefits of semi-supervised learning on unlabeled data.

\section{Background and Related Work}
Recently, \emph{TempLama} \cite{Dhingra2021TimeAwareLM} showed that directly placing the year of the timestamp as a prefix in the text is performative in the context of named-entity-recognition. They, then feed the date-prefixed inputs to a T5 \cite{Raffel2019ExploringTL} model to directly model the temporal context.
\citet{Cao2022TimeawarePF} directly compares a date-prefixing approach to an embedding approach where the date is numerically embedded with a linear projection. \cite{Cao2022TimeawarePF} in the context of text generation, found that linear projection was less sensitive to the timestamps while date-prefixing is better at generating more temporally sensitive facts.

Self-labeling (or self-distillation) is a semi-supervised learning strategy that typically involves learning from pseudo-labels for unlabeled data. Self-labeling is demonstrated to add performance gains across a variety of domains including text classification \cite{Shams2014SemisupervisedCF}.
\citet{Agarwal2021TemporalEO} found that self-labeling performs better than specialized pre-training objectives such as domain-adaptive pretraining \cite{Gururangan2020DontSP} across several tasks including sentiment analysis. However, it is important to note that recently \citet{Ushio2022NamedER} have shown that self-labeling, as presented in \cite{Agarwal2021TemporalEO}, is not as effective for NER when compared to models trained for specific time periods.

\section{Methodology}
\begin{figure}
  \centering
  \includegraphics[width=\linewidth]{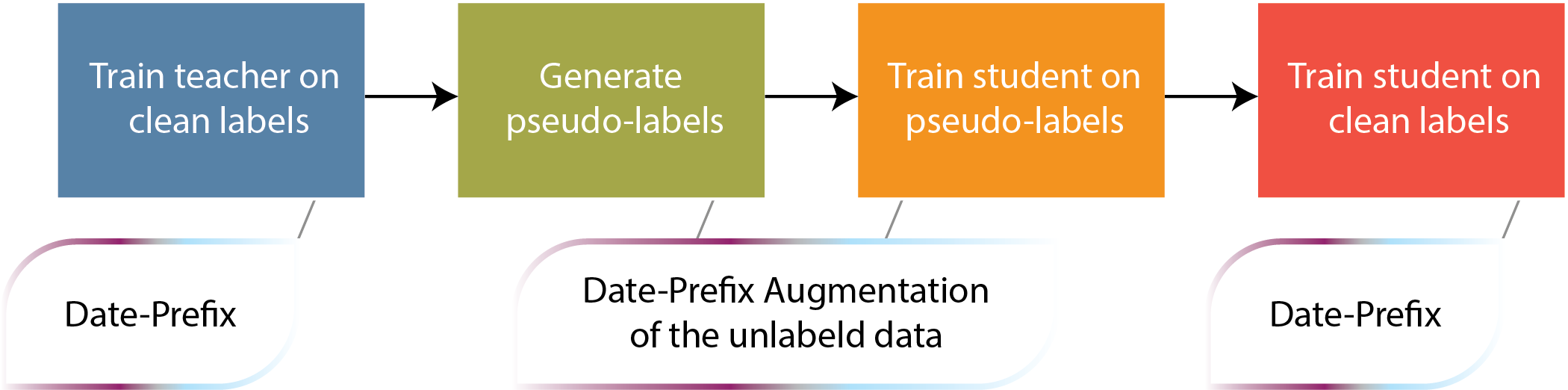}
  \caption{\textbf{Method Overview:} (top-row) summarization of our semi-supervised learning training pipeline stages, (bottom-row): modifications we made to the pipeline and at what stage they apply}
\label{fig:so}
\end{figure}
Figure \ref{fig:so} provides an overview of our system.
Following \citet{Agarwal2021TemporalEO}, we first train a teacher model on the full labeled dataset to create pseudo-labels for the unlabeled data. 
During this training phase, every sample in the labeled dataset is date-prefixed, meaning that the year of the timestamp is included as part of the input text.
We use a novel augmentation strategy on the date prefixes (see Section section \ref{sec:da}) to condition the pseudo-labels on the temporal context learned by the teacher. 
A new student model is then trained for 22000 training steps on the generated pseudo-labels and is subsequently trained on the original labeled data that was used for the teacher. Finally, we use the resulting student model for inference. For simplicity, both the teacher and student models share the same architecture. We provide further detail on the individual components of our system in the following sections.

\subsection{Pre-Trained Model}

Using a pre-trained language is generally much better than training a new model from scratch. However, it is not always clear which pre-training works best for any particular task. 
Here we use Bernice \cite{DeLucia2022BerniceAM} a variant of XLM-RoBERTa \cite{Conneau2019UnsupervisedCR} specialized for Twitter data.
We train a single model for inference on the test set and we do not rely on ensembling techniques.
We train using the cross-entropy classification loss.

\subsection{Date-Prefixing}
Consistent with \citet{Dhingra2021TimeAwareLM} we prefix each input text with the year of the given time-stamp followed by the text itself (e.g. “year: 2023 text: I really do enjoy drinks with friends”).
As we observe from Table \ref{tab:dp} training on this data conditions the model outputs with the temporal context found in the data using date-prefixing. Table \ref{tab:dp} provides real input and output examples based on a trained model across various years. We do not modify the architecture of the language model to take the timestamp as a vector input. By maintaining the use of textual-only input we are able to leverage any existing pre-trained models that have text-embedding only input.

\subsection{Date-Prefix Augmentation}
\label{sec:da}
When creating pseudo-labels to train a student model we use an augmentation strategy that takes advantage of our date-prefixing. Namely, given an unlabeled sample and its timestamp we randomly replace the year in the timestamp with a year between 2013 and 2021. Where, the years 2013 and 2021 are the earliest and latest years found in the labeled datasets, respectively. We perform an ablation experiment (see Section \ref{exp}) demonstrating that this augmentation strategy outperforms non-augmented self-labeling on the evaluation set.

\subsection{Training and Evaluation}
We use a single model trained using both the training and development sets for two epochs for inference on the test set. Model parameters using the Adam optimizer \cite{Kingma2014AdamAM} with a constant learning rate of 1e-5 using the binary-cross-entropy loss. Performance is measured using the macro-averaged F1 score of the future samples. 

\section{Experiments}
\label{exp}

\begin{table*}
  \caption{\textbf{Date Prompt Conditioning:} A demonstration of the date-prompting and subsequent model outputs conditioned on the prefix year. The model output is between 0 and 1, where the input is considered positive only if the output is above 0.5. The example input text is taken from the \emph{LongEval-Classification} dataset \cite{clef-longeval-alkhalifa-2023}.}
  \label{tab:dp}
  \begin{tabular}{ccccl}
    \toprule
    Example input & Output & Label & Orginal Year & Prefix Year\\
    \midrule
    "year: \textcolor{orange}{2013} text: I really do enjoy being single" & \textcolor{orange}{$0.503$} & positive & 2018 & \textcolor{orange}{2017}\\
    "year: \textcolor{purple}{2018}   text: I really do enjoy being single" & \textcolor{purple}{$0.510$}   & positive & 2018 & \textcolor{purple}{2018} \\
    "year: \textcolor{blue}{2023} text: I really do enjoy being single" & \textcolor{blue}{$0.495$} & negatuve & 2018 &  \textcolor{blue}{2023}\\
  \bottomrule
\end{tabular}
\end{table*}
\subsection{Experimental Setup}
In this section, we will compare the performance of models trained with and without the proposed augmentation strategies for pseudo-label generation. 
Namely, we will use a trained teacher model to generate labels with and without date-prefix augmentation. Subsequently, we a student models on each of the two sets of pseudo labels for 6000 training steps. Finally, then compare the downstream performance of each model.

Models will only be provided labels for the training set and trained until saturation on the interim evaluation set. 
For our experiments, we report the macro-averaged F1 scores for each subset of the evaluation set. We will also report the Relative Performance Drop (RPD) \cite{Alkhalifa2021OpinionsAM} for comparison between short and long-term time differences with respect to model performance.  

\begin{equation}
  \text{RPD}=\frac{f_{t_{j}}^{score} - f_{t_{0}}^{score}}{f_{t_{0}}^{score}}
\end{equation}

\subsection{Results}
We report the evaluation results of our experiments in Table \ref{tab:res}. 
Indeed, we see an overall improvement in performance especially when we observe the ‘short’ evaluation set results when using our full framework.
Additionally, the model using date-prefix augmentation gives by far the best RDP of $-0.0532$ with respect to the ‘within‘ and ‘short’ evaluation sets.
Note that the non-augmented models gives the best RDP of $-0.0411$ with respect to the ‘within‘ and ‘long‘ evaluation sets. However, when finetunning this same model on the gold labels, the RPD more than doubles to $-0.0852$ and is much worse than our full framework with $-0.0681$. 
A similar drop in performance can be seen when observing the F1 score on the ‘long’ evaluation set. It appears that fine-tuning the non-augmented model with clean data incurs a significant drop in performance. 
However, it is clear that our proposed augmentation strategy can leverage the older labeled data and attain significant performance gains. 

\section{Conclusion}
In this paper, we introduce a competitive framework for preserving the performance of sentiment analysis models across various temporal periods.
We promote date-prefixing, as a straightforward solution to condition the output of pre-trained language models with the temporal context of input text.
Furthermore, we build on the self-labeling framework developed by \citet{Agarwal2021TemporalEO}. Namely, given our date-prefix formatting, we can generate pseudo-labels conditioned on the temporal context of the input text.
We verify the performance gains of our proposed system against self-labeling without our augmentation strategy in our ablation experiments.
Altogether, our system yields competitive performance in overall score and attains the best RPD for the short evaluation set \cite{longevaloverview2023}.


\begin{table*}
  \caption{\textbf{Abalation Results}: 
 Results on the evaluation set, testing our self-labeling augmentation strategy. (\emph{baseline}: only using gold labels, \emph{+sl}: trained on pseudo-labels generated from \emph{baseline}, \emph{+ft}: fine-tuned on gold labels, \emph{(aug)}: date-prefix augmentation, \emph{(no-aug)}: no augmentation applied)
  We report the macro F1 score alongside the RPD between the various evaluation sets.
  The best results are highlighted}
  \label{tab:res}
  \begin{tabular}{clccccc}
    \toprule
    Method&F1 Within &F1 Short &F1 Long &RPD Within-Short &RPD Within-Long\\
    \midrule
    \emph{baseline} &0.7266 &0.6725 &0.6595 &-0.0744 &-0.0924\\
    
    \emph{+sl(no-aug)} &0.7213 &0.6747 &\textbf{0.6916} &-0.0646 &\textbf{-0.0411}\\
    \emph{+sl(no-aug)}+ft &\textbf{0.7355} &0.6728 &0.6728 &-0.0852 &-0.0852\\

    \emph{+sl(aug)} &0.7278 &0.6749 &0.6648 &-0.0727 &-0.0865\\
    \emph{+sl(aug)+ft (ours)} &0.7210 &\textbf{0.6833} &0.6719 &\textbf{-0.0532} &-0.0681\\

  \bottomrule
\end{tabular}
\end{table*}

\bibliography{sample-ceur}

\end{document}